\documentclass{ieeeaccess}
\usepackage{cite}
\usepackage{amsmath,amssymb,amsfonts}
\usepackage{algorithm}
\usepackage{algpseudocode}
\usepackage{graphicx}
\usepackage{textcomp}
\usepackage{hyperref}
\usepackage{tabularx}
\usepackage{caption}
\usepackage{subcaption}
\usepackage{float}
\usepackage[super]{nth}
\usepackage[font=small,format=plain,labelfont={bf,up},labelsep=period]{caption}

\DeclareCaptionLabelFormat{table}{TABLE~#2}
\renewcommand\thetable{\Roman{table}}
\def\BibTeX{{\rm B\kern-.05em{\sc i\kern-.025em b}\kern-.08em
    T\kern-.1667em\lower.7ex\hbox{E}\kern-.125emX}}
\begin{document}
\history{Date of publication xxxx 00, 0000, date of current version xxxx 00, 0000.}
\doi{10.1109/ACCESS.2017.DOI}

\title{Novel Epileptic Seizure Detection Techniques and their Empirical Analysis}
\author{\uppercase{Rabel Guharoy}\authorrefmark{1}, \IEEEmembership{Member, IEEE},
\uppercase{Nanda Dulal Jana}\authorrefmark{2}, \uppercase{Suparna Biswas}\authorrefmark{3}, \uppercase{Lalit Garg}\authorrefmark{4}
}
\address[1]{School of Information Technology, Artificial Intelligence and Cyber Security, Rashtriya Raksha University, Gandhinagar, 382355, India}
\address[2]{Dept. of Computer Science and Engineering, National Institute of Technology, Durgapur, 713209, India}
\address[3]{Dept. of Electronics and Communication Engineering, Guru Nanak Institute of Technology, Kolkata, 700114, India}
\address[4]{Dept. of Information and Communication Technology, University of Malta, Msida, 2080, Malta}

\markboth
{R Guharoy \headeretal: Empirical Analysis of Various Epileptic Seizure Detection Techniques}
{R Guharoy \headeretal: Empirical Analysis of Various Epileptic Seizure Detection Techniques}

\corresp{Corresponding author: Rabel Guharoy (e-mail: rabelrock@gmail.com), Lalit Garg (email: lalit.garg@um.edu.mt)}

\begin{abstract}
An Electroencephalogram (EEG) is a non-invasive exam that records the brain's electrical activity. This is used to help diagnose conditions such as different brain problems. EEG signals are taken for epilepsy detection, and with Discrete Wavelet Transform (DWT) and machine learning classifier, they perform epilepsy detection. In Epilepsy seizure detection, machine learning classifiers and statistical features are mainly used. The hidden information in the EEG signal helps detect diseases affecting the brain. Sometimes it is complicated to identify the minimum changes in the EEG in the time and frequency domain's purpose. The DWT can give a suitable decomposition of the signals in different frequency bands and feature extraction. We use the tri-dimensionality reduction algorithm, Principal Component Analysis (PCA), Independent Component Analysis (ICA), and Linear Discriminant Analysis (LDA). Finally, features are selected by using a fusion rule and at the last step, three different classifiers, Support Vector Machine (SVM), Naive Bayes (NB), and K-Nearest-Neighbor (KNN) have been used individually for the classification. By Leveraging the combination of LDA and NB on the Bonn dataset, We have achieved remarkable score of $100\%$ across all evaluation metrics. These results outperform other classifier combinations, including $89.17\%$ for LDA and SVM, $80.42\%$ for LDA and KNN, $89.92\%$ for PCA and NB, $85.58\%$ PCA and SVM, $80.42\%$ PCA and KNN, $82.33\%$ for ICA and NB, $90.42\%$ for ICA and SVM, $90\%$ for ICA and KNN. The results prove the effectiveness of the LDA and NB combination model.
\end{abstract}


\titlepgskip=-15pt

\maketitle

\section{Introduction} \label{sec:introduction}

EPILEPSY is an abnormal electrical brain activity called a seizure, like an electrical storm inside the brain. A chronic neurological disorder is epilepsy. Inside the brain, electrical activity disturbance is the leading cause \cite{al2020general, world2006neurological, alarcon2012introduction, qu1993improvement}. 

It could be caused by different reasons \cite{sazgar2019absolute}, such as low sugar levels and a shortage of blood oxygen during childbirth \cite{sirven2015epilepsy, delanty1998medical}. Epilepsy affects approximately $50$ million people worldwide, with $100$ million suffering at least once in their lifetime \cite{sirven2015epilepsy}, \cite{caplan2017adhd}. Overall, epilepsy is responsible for $0.5\%$ of disease burden worldwide, and the control rate is $0.5\%-1\%$ \cite{sazgar2019absolute, sirven2014sticks}. Brain neurons detect it by analyzing the brain signals. Generate the signals through the neuron's connection with each neuron in a problematic way to share with human organs. Electroencephalogram (EEG) and Electrocorticography (ECoG) media monitor the same brain signal. These signals are complex, non-linear, non-stationary, noisy, and produce big data. Seizure detection and discovery is challenging work for brain-related knowledge. Classify EEG data and detect seizures with sensible patterns without compromising performance through the machine learning classifiers. The main challenge is selecting better classifiers and features. The last few years mainly focused on the machine learning classifiers and taxonomy of statistical features- 'black-box' and 'non-black-box'. Primarily focusing on state-of-the-art methods and ideas will give a better understanding of seizure detection and classification \cite{siddiqui2020review}. 

As per neuro experts, seizures can be divided into two types based on symptoms: partial and generalized \cite{caplan2017adhd, thurman2011standards}. DWT has been used for different groups of epilepsy types of EEG signals. DWT is better for the feature extraction step because it efficiently works in this field. Feature selection is used to minimize dimensionality without irrelevant features. It is used for differential evolution purposes. This research considers feature extraction and selection using efficient models for EEG classification. Seven varieties of wavelets were tested. Few kinds of DWT are used to process a spacious difference of features. In raw data, multi-level DWT and several sub-bands extract the features. Seven statistical functions were applied in features. These functions include Standard deviation (SD), Average power (AVP), Mean absolute value (MAV), mean, variance, Shannon entropy, and skewness. Choose better features to use these function values as an input to DE to classify the signals, using three matching metrics, six supervised machine learning and two ensemble learning methods. \cite{al2020general} analyzed various methods to classify cases of epilepsy by brain signals EEG using DE with DWT. The result is provided against various performance metrics, including accuracy, recall measures, and precision. \cite{al2020general} found Support Vector Machine (SVM) better for accuracy, and Naive Bayes (NB) and K-nearest Neighbours (KNN) better in convergent results. The EEG signals are non-linear, non-stationary, weak and time-varying. The process of EEG signal acquisition introduces some common noises like electrooculogram (EOG) and electrocardiogram (ECG) artefacts \cite{geng2022improved}. EEG signals can be proceeded by combining independent component analysis (ICA), common spatial pattern (CSP), and wavelet transform (WT) \cite{geng2022improved}. The ICA algorithm breaks the EEG signals into independent components. Then, these components are decomposed by WT to reach the wavelet coefficient of each independent source. The two-compromise threshold function is used to activate the wavelet packet coefficients. Then, the CSP algorithm extracts the denoised EEG data features. Lastly, four classification algorithms are used for feature classification. The result is better identified, removes EOG and ECG artefacts from the data, and preserves neural activity \cite{geng2022improved}. 

The literature review shows that most existing works have high time complexity due to high dimensional feature space. So, in this work, we have tried to reduce the feature dimension in two steps. In the first step after applying DWT, three different dimensionality reduction techniques are used to reduce the feature dimension. A feature-level fusion technique has been used in the next step for further feature dimension reduction. Finally, in the last step, three different classifiers have been used to detect epilepsy with high accuracy. The rest of the paper is organized as follows: Section \ref{sec:literature} provides a brief review of the epilepsy detection method, Section \ref{sec:background} provides all the techniques, Section \ref{sec:proposedmethodology} provides our proposed methodology and algorithm, and Section \ref{sec:results} provides the results and discussion. Finally, we conclude the paper in Section \ref{sec:conclusion}

\section{Literature review} \label{sec:literature}

This section summarizes the previous research contributions to epilepsy prediction techniques. According to A. Prochazka et al. \cite{prochazka2008wavelet} and S. Cinar et al. \cite{ccinar2017novel}, numerous machine learning applications are seen in health and biological data sets for better outcomes. Researchers/scientists in different areas, specifically data mining and machine learning, are actively involved in proposing solutions for better seizure detection. Machine learning has been significantly applied to discover sensible and meaningful patterns from domain datasets \cite{fayyad1996advances, yin2011data}. It plays a significant and potential role in solving the problems of various disciplines like healthcare \cite{fayyad1996advances, islam2016brand, aljumah2013application, aljumah2016data, siddiqui2020correlation, aljumah2014hypertension, almazyad2010effective, singh2019performance}. Machine learning applications can also be seen on brain datasets for seizure detection, epilepsy lateralization, differentiating seizure states, and localization \cite{fayyad1996advances, yin2011data, islam2016brand}. This has been done by various machine learning classifiers such as ANN, SVM, decision trees, decision forests, and random forests \cite{fayyad1996advances, yin2011data, islam2016brand, aljumah2013application}. Amin et al. \cite{amin2015feature} proposed a DWT-based feature extraction scheme for classifying EEG signals. DWT was applied to EEG signals, and the relative wavelet energy was calculated from the last decomposition level's detailed coefficients and approximation coefficients. The EEG dataset used in \cite{amin2015feature} consisted of two classes: (1) EEG recorded during Raven's advanced progressive metric test and (2) EEG recorded in resting condition with eyes open. An accuracy of 98\% was obtained in \cite{amin2015feature} by using the SVM with approximation (A4) and detailed coefficients (D4). It was observed that their feature extraction approach had the potential to classify the EEG signals recorded during a complex cognitive task and also achieved a high accuracy rate. 

Al-Qerem et al. \cite{al2020general} developed a Wavelet family, and differential evolution is proposed for categorizing epilepsy cases based on EEG signals. DWT is widely used in feature extraction due to its efficiency, as confirmed by the results of previous studies \cite{al2020general}. The feature selection step is used to minimize dimensionality by excluding irrelevant features. This step is conducted using differential evolution. \cite{al2020general} presents an efficient model for EEG classification by considering feature extraction and selection. \cite{al2020general} tested seven different types of standard wavelets. These are Discrete Meyer (dmey), Reverse biorthogonal (rbio), Biorthogonal (bior), Daubechies (db), Symlets (sym), Coiflets (coif), and Haar (Haar). Different types of feature extraction transform different types of discrete wavelets. \cite{al2020general} uses differential evolution to choose appropriate features that will achieve the best performance of signal classification. \cite{al2020general} used Bonn databases to build the classifiers and test their performance for the classification step. The results prove the effectiveness of the proposed model. Epilepsy is a severe chronic neurological disorder detected by analyzing the brain neurons' signals. Neurons are connected in a complex way to communicate with human organs and generate signals.

Monitoring these brain signals is commonly done using EEG and Electrocorticography (ECoG) media. These signals are complex, noisy, non-linear, and non-stationary and produce a high volume of data. Hence, the detection of seizures and the discovery of brain-related knowledge is a challenging task. Machine learning classifiers can classify EEG data, detect seizures, and reveal relevant sensible patterns without compromising performance. As such, various researchers have developed several approaches to seizure detection using machine learning classifiers and statistical features. M. K. Siddiqui et al. \cite{siddiqui2020review} suggested the main challenges are selecting appropriate classifiers and features. \cite{siddiqui2020review} aims to present an overview of the wide varieties of these techniques over the last few years based on the taxonomy of statistical features and machine learning classifiers—'black-box' and 'non-black-box'. The presented state-of-the-art methods and ideas will give a detailed understanding of seizure detection, classification, and future research directions. ECG is the P-QRS-T wave, representing the cardiac function \cite{martis2013ecg}. The information concealed in the ECG signal is useful in detecting the disease afflicting the heart. Identifying the subtle changes in the ECG in time and frequency domains is very difficult. DWT can provide reasonable time and frequency resolutions and decipher the hidden complexities in the ECG. In \cite{martis2013ecg}, five types of beat classes of arrhythmia as recommended by the Association for Advancement of Medical Instrumentation (AAMI) were analyzed, namely: non-ectopic beats, supra-ventricular ectopic beats, ventricular ectopic beats, fusion betas, and unclassifiable and paced beats. Three dimensionality reduction algorithms, Principal Component Analysis (PCA), Linear Discriminant Analysis (LDA) and Independent Component Analysis (ICA), were independently applied to DWT subbands for dimensionality reduction. These dimensionality-reduced features were fed to the SVM, neural network (NN), and probabilistic neural network (PNN) classifiers for automated diagnosis. In combination with PNN with a spread value of $0.03$, the ICA features performed better than the PCA and LDA. Using a ten-fold cross-validation scheme, \cite{martis2013ecg} yielded an average sensitivity, specificity, positive predictive value (PPV) and accuracy of$ 99.97\%$, $99.83\%$, $99.21\%$ and $99.28\%$, respectively.

B. Shi et al. \cite{shi2021binary} proposed a new binary harmony search (BHS) to select the optimal channel sets and optimize the system's accuracy. The BHS is implemented on the training data sets to determine the optimal channels, and the test data sets are used to evaluate the classification performance on the selected channels. The sparse representation-based classification, LDA, and SVM are performed on the CSP features for motor imagery (MI) classification. Two public EEG datasets are employed to validate the proposed BHS method. The paired t-test is conducted on the test classification performance between the BHS and traditional CSP with all channels. The results reveal that the proposed BHS method significantly improved classification accuracy compared to the conventional CSP method ($p < 0.05$). It plays a significant and potential role in solving the problems of various disciplines like healthcare \cite{fayyad1996advances}. The EMD algorithm decomposes the EEG signal into a finite number of intrinsic mode functions (IMFs) with a $3$-fold cross-validation method to get $70.72\%$ mean sensitivity and $95.37\%$ mean specificity \cite{agrawal2013application}. EMD algorithm and a machine learning–based classifier that is robust enough for practical application purposes and results in $100\%$ accuracy, specificity of $95.5\%$, and latency of $2.53$ seconds \cite{agrawal2019early}. One fast method of data acquisition, feature extraction and feature space creation for epileptic seizure detection and get positive results reaching up to $99.48\%$ Sensitivity \cite{bugeja2016novel}. The predictability is discussed regarding the latency and the required data length for the proposed approach over the state-of-the-art method in EEG-based seizure prediction \cite{bonello2018effective}. The EEG signals are weak, non-linear, non-stationary, and time-varying. Hence, a practical feature extraction method is the key to improving recognition accuracy. EOG and electrocardiogram artefacts are common noises in the process of EEG signal acquisition, which seriously affects the extraction of useful information. 

X. Geng et al. \cite{geng2022improved} propose a processing method for EEG signals by combining independent component analysis (ICA), WTs, and CSP. First, the ICA algorithm breaks the EEG signals into independent components. Then, these independent components are decomposed by WT to obtain the wavelet coefficient of each independent source. The soft and hard compromise threshold function processes the wavelet packet coefficients. Then, the CSP algorithm is used to extract the features of the denoised EEG data. Finally, four common classification algorithms verify the improved algorithm's effectiveness. The experimental results show that the EEG signals processed by the proposed method have obvious advantages in identifying and removing electrooculogram (EOG) and ECG artefacts; meanwhile, it can preserve the neural activity missed in the noise component. Cross-comparison experiments also proved that the proposed method has higher classification accuracy than other algorithms. Kapoor B et al. \cite{kapoor2022epileptic} proposed a hybrid classifier for epileptic seizure prediction using the AdaBoost classifier, random forest classifier, and the decision tree classifier and achieved $96.6120\%$ accuracy and $91.3684\%$ specificity with CHB-MIT data set and $95.3090\%$ accuracy, $90.0654\%$ specificity with Siena Scalp data. Abdulhamit Subasi et al. \cite{amin2015feature} investigated EEG signal classification using PCA, ICA, LDA + Support Vector Machines and achieved $100\%$ accuracy and $100\%$ specificity (LDA) with the Bonn dataset. Ping Tan et al. \cite{tan2020dimensionality} explored dimensionality reduction for feature selection in BCI using the Bonn dataset and DimReM-NMBDE with SVM, KNN and DA, which provided an accuracy of $95.00\%$, $93.57\%$, and $94.29\%$, respectively. S. Priyanka et al. \cite{priyanka2017feature} also used the Bonn dataset with the Artificial Neural Networks and got $96.9\%$ accuracy.

\section{Background} \label{sec:background}

This section describes all the tools and techniques used in this work. First, we have explained DWT's importance and function; in the following subsection, all the dimensionality techniques, such as PCA, ICA, and LDA, are explained. 

\subsection{Discrete Wavelet Transform (DWT)} \label{subsec:dwt}

Frequency domain techniques are prevalent feature extraction techniques for different classification problems. WT is one of the frequency-based feature extraction approaches that shows the property of time-frequency localization [WT1] [WT2] and is suitable for analyzing non-stationary signals. ECG signals are non-stationary. WT is an effective tool for analyzing ECG signals \cite{agrawal2013application, agrawal2019early}.

\begin{figure}[!htb]
    \centering
    \includegraphics[width=0.48\textwidth]{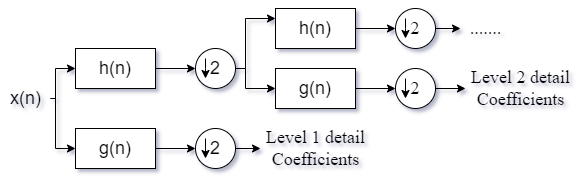}
    \caption{\nth{2} level of Coefficient}
    \label{fig:fig1}
\end{figure}

DWT decomposes a one-dimensional signal into two sub-bands: high and low-frequency. This high-frequency sub-band is called detail, and the low-frequency sub-band is called approximation. Let $x$ be a one-dimensional signal. In the case of DWT's case, these samples are first passed through a low-pass filter with impulse $g$ and a high-pass filter $h$. Fig. \ref{fig:fig1} can be described using mathematical formulations shown in equation \ref{equation1}

\begin{equation} \label{equation1}
    y[n] = (x * g) [n] = \sum_{k = -\infty}^\infty x[k] g[n - k]
\end{equation}

This equation result and signal are decomposed and continue using a high-pass filter $h$. The outputs give the detail coefficients from the high-pass filter and approximation coefficients from the low-pass filter $g$. These two filters are related, and the DWT decomposition of a signal using the filter band is shown in Fig \ref{fig:fig1}

This signal passes through the filter $h(n)$ and down factor of $2$ to obtain the approximation coefficient in the first level. Following the same process, a signal is passed through another filter, $g(n)$ and a down sample of the same factor of $2$ obtains the detail coefficients. In the second level, approximation coefficients are passed through the same filter, $h(n)$ and $g(n)$, and the downsample obtains the coefficients in the second level. 
i.e.,

\begin{equation} \label{equation2}
    y\textsubscript{low}[n] = \sum_{k=-\infty}^\infty x[k]g[2n-k]
\end{equation}

\begin{equation} \label{equation3}
    y\textsubscript{high}[n] = \sum_{k=-\infty}^\infty x[k]h[2n-k] 
\end{equation}

We get output from equations \ref{equation2}
and \ref{equation3}, which is half the signal's frequency band. So, each result has half the frequency band of the input. After that, frequency resolution has been doubled. 

i.e., subsampling operator $\downarrow$

\begin{equation} \label{equation4}
    (y \downarrow k) [n] = y [kn]
\end{equation}

The above sum can be written more constructively,

\begin{equation} \label{equation5}
     y\textsubscript{low} = (x * g) \downarrow 2
\end{equation}

\begin{equation} \label{equation6}
     y\textsubscript{high} = (x * h) \downarrow 2
\end{equation}

\subsection{Principal Component Analysis (PCA)} \label{subsec:pca}

PCA is a linear dimensionality reduction method that projects the data towards the direction of maximum variableness \cite{salem2019data}. It is widely used to reduce the dimensions of large data sets within a small one that contains most of the information from the large data set. It is also used for improving the performance of different machine-learning algorithms. The method works by representing data in a space that best represents the variation in terms of sum-squared error. The method is also helpful in segmenting signals from different sources.

Step $1$: Compute the covariance matrix from the data as,

\begin{equation} \label{equation7}
    C = (X - \bar{x}) (X - \bar{x}) ^ T
\end{equation}

where $X$ is the data matrix of DWT coefficients in a subband of $N \times 100$ dimension, $N$ is the total number of patterns, and $\bar{x}$ represents the mean vector of $X$.

Step $2$: Compute the matrix of eigenvectors $V$ and diagonal matrix of eigenvalues $D$ as

\begin{equation} \label{equation8}
    V\textsuperscript{-1}CV = D
\end{equation}

Step $3$: The eigenvectors in $V$ are sorted in the descending order of eigenvalues in $D$, and the data is projected on these eigenvector directions by taking the inner product between the data matrix and sorted eigenvector matrix as,

\begin{equation} \label{equation9}
    \text{Projected data} = [V^T (X - \bar{x})^T]^T
\end{equation}

\subsection{Independent Component Analysis (ICA)} \label{subsec:ica}

ICA can be applied to mixed signals. Independence relates to the ability to guess one component from the information carried by others. Statistically, it denotes that the joint probability of independent quantities is earned due to the multiplication of the probability value of each component. The ICA algorithm is used for statistical techniques that may be effective in areas that remove noise and EEG signals. Previously, this ICA-based technique was used to appropriate and remove pollution currents in brain wave planning \cite{yin2011data, islam2016brand}. Mainly, a search technique focused on minimizing the effect of surrounding parameters in EEG signal form. 

If $x$ is a vector with mixtures ${x\textsubscript{1}, x\textsubscript{2}, . . . , x\textsubscript{n}}$ and let $s$ be the source vector with ${s\textsubscript{1}, s\textsubscript{2}, . . . , s\textsubscript{n}}$. 

Let$ A$ denote the weight matrix with elements $a\textsubscript{ij}$. The ICA model assumes that the signal $x$ (the DWT coefficients in a subband) we observed is linearly mixed with the source signals. The ICA model is given by

\begin{equation} \label{equation10}
    x = As\ or\ x = \sum_{i=1}^{n} a\textsubscript{i}s\textsubscript{i}
\end{equation}

Equation \ref{equation10} is called ICA. The problem is determining the matrix $A$ and the independent components $s$, knowing only the measured variables $x$. The only assumption the methods take is that the components are independent. It has also been proved that the components must have a non-gaussian distribution \cite{siddiqui2020review}. ICA looks a lot like the "blind source separation" (BSS) problem or blind signal separation: a source is an original signal in the ICA problem, an independent component. There is also no information about the independent components in the ICA case, like in the BSS problem \cite{jain2012blind}. Whitening can be performed via eigenvalue decomposition of the covariance matrix:

\begin{equation} \label{equation11}
    VDV^T = E[\hat{x}\hat{x}^T]
\end{equation}

where $V$ is the matrix of orthogonal eigenvectors, and $D$ is a diagonal matrix with the corresponding eigenvalues. The whitening is done by multiplication with the transformation matrix $P$:

\begin{equation} \label{equation12}
    \Tilde{x} = P\Tilde{x}
\end{equation}

\begin{equation} \label{equation13}
    P = VD\textsuperscript{1/2}V^T
\end{equation}

The matrix for extracting the independent components from $\Tilde{x}$ is $\Tilde{W}$, where $P = \Tilde{W}P$

\subsection{Linear Discriminant Analysis (LDA)} \label{subsec:lda}

The LDA technique finds a linear combination of features that separates or characterizes two or more classes of objects or events. LDA aims to find a feature subspace that maximally separates the groups. LDA generates a new variable, which joins the original predictors. This is achieved by maximizing the differences between the predefined groups concerning the new variable. The predictor scores are combined to obtain a new discriminant score. It can also be visualized as a data dimension reduction method with a one-dimensional line for $p$-dimensional predictors. Mainly based on the linear score function, a function of a class, $\mu\textsubscript{i}$, and the pooled variance-covariance matrix. The Linear Score Function is defined as:

\begin{align} \label{equation14}
    s_i^L(X) &= -\frac{1}{2} \mu_i \Sigma^{-1} \mu_i + \mu_i \Sigma^{-1} x + \log P(\pi_i) \nonumber \\
    &= d_{i0} + \sum_{j=1}^{p} d_{ij}x_j + \log P(\pi_i) \nonumber \\
    &= d_i^L(X) + \log P(\pi_i)
\end{align}

where $d_{i0} = -\frac{1}{2} \mu_i \Sigma^{-1} \mu_i$ and $d_{ij} = j\textsuperscript{th}$ element of $\mu_i \Sigma^{-1} \mu_i$ and we call $d_i^L(X)$ the linear discrimination function. As we can see from equation \ref{equation14}, the far right-hand expression is similar to linear regression with intercept $d_{i0}$ and coefficients $d_{ij}$.

\subsection{Contributions}

The main contributions of this  work are  presented as follows:

\begin{itemize}
    \item In this paper we have presented three different methods of epilepsy detection techniques by combining the 5th level of DWT, three different dimensionality reduction techniques PCA, LDA, and ICA and at the last step three different classifiers SVM, KNN and NB.
    \item  Most of the existing work has used a large dimension of feature space which indirectly increases the time complexity. But in this work, we have integrated feature extraction with feature selection to reduce the dimension of features.
    \item The proposed method can detect epilepsy with very high accuracy.
\end{itemize}

\section{Proposed Methodology} \label{sec:proposedmethodology}

\begin{figure}[!ht]
    \centering
    \includegraphics[width=0.46\textwidth]{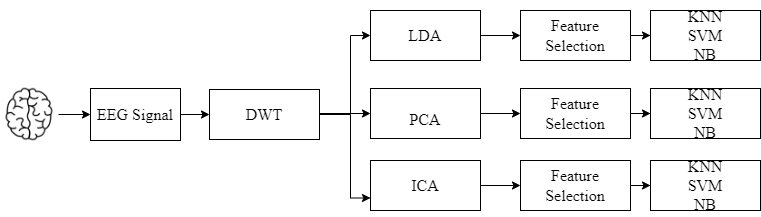}
    \caption{Block Diagram of the proposed method.}
    \label{fig:fig2}
\end{figure}

\begin{figure}[!ht]
    \centering
    \includegraphics[width=0.48\textwidth]{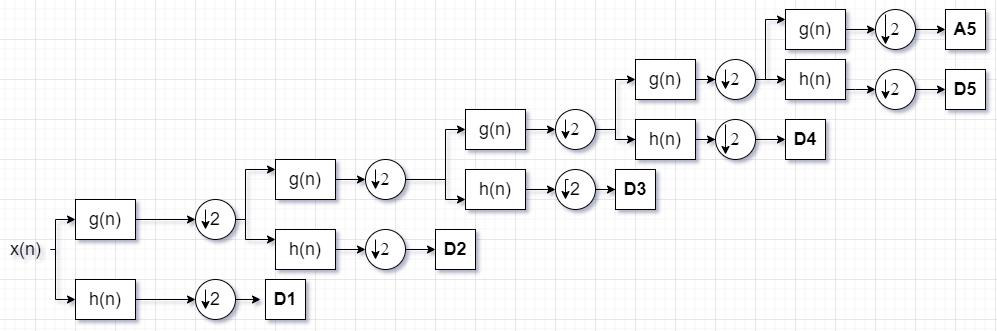}
    \caption{Block Diagram of 5\textsuperscript{th} level Decomposition of EEG signal}
    \label{fig:fig3}
\end{figure}

A schematic overview of our proposed methodology is depicted in Fig \ref{fig:fig2}, explaining the working of each and every block. In the present work, the EEG signal is decomposed by DWT up to level five using Daubechies (dB1), as shown in Fig \ref{fig:fig3}

\begin{figure}[!ht]
    \centering
    \includegraphics[width=0.48\textwidth]{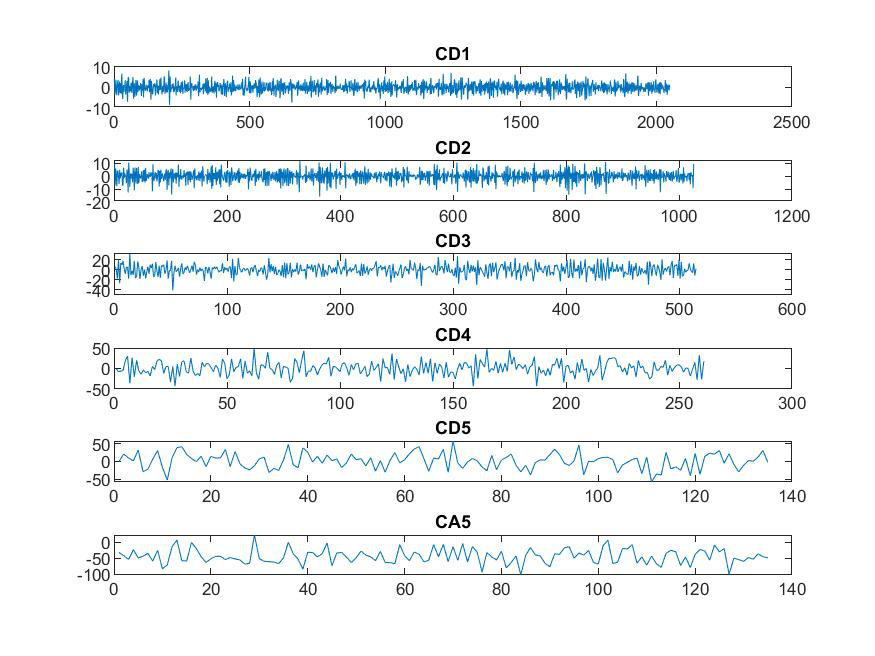}
    \caption{Block Diagram of 5\textsuperscript{th} level Decomposition of EEG signal}
    \label{fig:fig4}
\end{figure}

After the $5\textsuperscript{th}$ level decomposition, we worked with six subbands which are CD1($1\textsuperscript{st}$ level detail coefficient), CD2($2\textsuperscript{nd}$ level detail coefficient), CD3($3\textsuperscript{rd}$ level detail coefficient), CD4($4\textsuperscript{th}$ level detail coefficient), CD5($5\textsuperscript{th}$ level detail coefficient), CA5($5\textsuperscript{th}$ level approximation coefficient). All these six subbands for one sample of ECG signal are shown in Fig \ref{fig:fig4}.

After the 5\textsuperscript{th} level wavelet decomposition, we applied three different dimensionality reduction techniques, PCA, LDA and ICA, on the six selected subbands by setting the dimension of features of length L. First, we selected anyone dimensionality reduction technique and applied it on approximation coefficients (CA5) and detail coefficients CD1, CD2, CD3, CD4, CD5 and extracted the features CA5\_L, CD1\_L, CD2\_L, CD3\_L, CD4\_L, CD5\_L of length L.

Then, a feature-level fusion technique has been applied to further dimension reduction. The whole algorithm is explained in Algorithm \ref{alg:alg1}. 

\begin{algorithm}
\caption{}\label{alg:alg1}
\begin{flushleft}
\begin{algorithmic}[1]
    \State Select dimensionality reduction technique PCA/LDA/ICA
    \State Set dimension of features (L)
    \State Apply dimensionality reduction technique on approximation coefficients (CA5) and extract the features CD5\_L
    \State Apply dimensionality reduction technique on detail coefficients CD1, CD2, CD3, CD4, CD5 and extract the features $CD1\_L$, $CD2\_L$, $CD3\_L$, $CD4\_L$, $CD5\_L$
    \State Compute $CD_L = \text{MAX}(CD1\_L, CD2\_L, CD3\_L, CD4\_L, CD5\_L)$
    \State Compute $F = \mu_1 \cdot CA5\_L + \mu_2 \cdot CD_L$
    \State Select any one classifier (KNN or SVM, or NB)
    \State Obtain the class.
\end{algorithmic}
\end{flushleft}
\end{algorithm}

For the feature selection, we used the max function and then the linear combination rule described in Step $5$ of Algorithm \ref{alg:alg1}. Finally, combine the features by applying the fusion rule discussed in step $6$. Here, the value of $\mu\textsubscript{1}$ and$ \mu\textsubscript{2}$ has been selected by using the trial-and-error method, following the linear combination rule $\mu\textsubscript{1}+ \mu\textsubscript{2}=1$. For this method, $\mu\textsubscript{1}$ and $\mu\textsubscript{2}$ are set as $0.7$ and $0.3$. In classification, a dataset A to E has a set of 'class attributes' and a 'non-class attribute'. They are the principal components, and their pertinent knowledge is most important, as both strongly associate with potential classification. The target attribute is the 'class attribute' C, comprising more than one class value, e.g., seizure and non-seizure. On the contrary, attributes $A = {A\textsubscript{1}, A\textsubscript{2}, A\textsubscript{3} . . . A\textsubscript{n}}$ are known as 'non-class attributes' or predictors \cite{fayyad1996advances, adnan2016forest}. The following classifiers have been popularly used in seizure detection. Common classifiers such as SVM \cite{cortes1995support}, decision tree \cite{quinlan2014c4} and decision forest \cite{breiman2001random} are applied to the processed EEG dataset for seizure detection. We are using three different classifier algorithms. In the algorithm, we take parameters: gradient, solver, bin edges, alpha, beta, bias, Mu, sigma, etc. We get better results after the classifier algorithm. Finally, three different classifiers, SVM, NB and KNN, were applied for classification at the last stage. The whole feature extraction and feature selection algorithm is explained in Algorithm \ref{alg:alg1}.

\section{Result and Discussion} \label{sec:results}

\textbf{Data sets}: In this work, we have used the Bonn dataset, which was recorded at Bonn University. This dataset is widely used for the detection of epilepsy \cite{li2017epileptic}. This dataset is publicly available and sampled at $173.6$Hz with a $23.6$s duration. It consists of $500$ EEG signals of five classes named S, F, N, O, and Z. Each category has $100$ different EEG signals \cite{eeg_databases}. This recorded data is considered the highest accuracy. All signals are recorded in the same $128$-channel amplifier system channel. Each section has a different acquisition of circumstances, like open eyes, closed eyes, seizure-free status, seizure-free status, inside five epileptogenic zones and seizure activity. The Bonn dataset details are explained in Table \ref{tab:tab1}.

\begin{table}[ht]
    \centering
    \begin{tabularx}{0.48\textwidth}{|c|c|c|X|}
        \hline
        \textbf{Set Name} & \textbf{Annotation of Data} & \textbf{Size} &  \textbf{Acquisition Circumstances} \\
        \hline
        Set A & Z000.txt—Z100.txt & 564 KB & Five healthy subjects with open eyes \\
        \hline
        Set B & O000.txt—O100.txt & 611 KB & Five healthy subjects with closed eyes \\
        \hline
        Set C & N000.txt—N100.txt & 560 KB & Five people with epilepsy with seizure-free status \\
        \hline
        Set D & F000.txt—F100.txt & 569 KB & Five people with epilepsy with seizure-free status inside five epileptogenic zones \\
        \hline
        Set E & S000.txt—S100.txt & 747 KB & Five subjects during seizure activity \\
        \hline
    \end{tabularx}
    \captionsetup{justification=centering}
    
    \caption{Samples of data in normal and seizure cases}
    \label{tab:tab1}
\end{table}

We have applied a $10$-fold cross-validation technique for partitioning training and testing classifiers. Tables \ref{tab:tab2}, \ref{tab:tab3} and \ref{tab:tab4} represent the various performance measures for the proposed method by selecting the dimensionality reduction technique ICA with KNN, SVM and NB. These tables are focused on the relative similarity between the various performance measures. As per table data, we have achieved the maximum average accuracy of $100\%$ for combining ICA with NB in the $2$-class classification between the A-E dataset. We have also computed other performance measures like F-measure, Recall, Specificity, sensitivity and precision, which are given in Tables \ref{tab:tab2}, \ref{tab:tab3} and \ref{tab:tab4}. The proposed method achieved the maximum average sensitivity of $100\%$ for all three combinations (ICA+KNN), (ICA+NB) and (ICA+SVM) in the case of the A-E data set. From Table \ref{tab:tab3}, it is also observed that we have achieved the maximum average sensitivity of $100\%$ for the dataset B-E. The comparison of Tables \ref{tab:tab2}, \ref{tab:tab3} and \ref{tab:tab4} shows that the maximum average Specificity and F-measure value is obtained for the A-E dataset. 

\begin{table}[!htb]
    \centering
    \setlength{\tabcolsep}{0.157\tabcolsep}
    \begin{tabular*}{\columnwidth}{|c|*{6}{>{\centering\arraybackslash}p{1.2cm}|}}
        \hline
        \textbf{CASE} & \textbf{Accuracy (\%)} & \textbf{Sensitivity (\%)} & \textbf{Specificity (\%)} & \textbf{Precision (\%)} & \textbf{Recall (\%)} & \textbf{F-measure} \\
        \hline
        A-C & 88.50 & 93.99 & 81.14 & 85.24 & 93.99 & 0.89 \\
        \hline
        A-D & 83.50 & 82.65 & 83.87 & 83.99 & 82.65 & 0.82 \\
        \hline
        A-E & 93.00 & 100.00 & 86.01 & 88.36 & 100.00 & 0.94 \\
        \hline
        B-C & 91.50 & 93.33 & 89.83 & 90.47 & 93.33 & 0.92 \\
        \hline
        B-D & 91.50 & 93.31 & 90.37 & 90.12 & 93.31 & 0.91 \\
        \hline
        B-E & 92.00 & 100.00 & 84.07 & 86.32 & 100.00 & 0.92 \\
        \hline
    \end{tabular*}
    \captionsetup{justification=centering}
    
    \caption{Results of the proposed model for the combination of ICA with KNN.}
    \label{tab:tab2}
\end{table}

\begin{table}[!htb]
    \centering
    \setlength{\tabcolsep}{0.157\tabcolsep}
    \begin{tabular*}{\columnwidth}{|c|*{6}{>{\centering\arraybackslash}p{1.2cm}|}}
        \hline
        \textbf{CASE} & \textbf{Accuracy (\%)} & \textbf{Sensitivity (\%)} & \textbf{Specificity (\%)} & \textbf{Precision (\%)} & \textbf{Recall (\%)} & \textbf{F-measure} \\
        \hline
        A-C & 88.00 & 92.68 & 83.67 & 86.33 & 92.68 & 0.89 \\
        \hline
        A-D & 85.50 & 96.03 & 75.96 & 79.14 & 96.03 & 0.86 \\
        \hline
        A-E & 97.50 & 100.00 & 94.96 & 95.55 & 100.00 & 0.98 \\
        \hline
        B-C & 86.50 & 83.63 & 89.65 & 91.07 & 83.63 & 0.87 \\
        \hline
        B-D & 90.50 & 89.63 & 92.59 & 91.72 & 89.63 & 0.90 \\
        \hline
        B-E & 94.50 & 100.00 & 88.78 & 90.69 & 100.00 & 0.95 \\
        \hline
    \end{tabular*}
    \captionsetup{justification=centering}
    
    \caption{Results of the proposed model for the combination of ICA with SVM}
    \label{tab:tab3}
\end{table}

\begin{table}[!htb]
    \centering
    \setlength{\tabcolsep}{0.157\tabcolsep}
    \begin{tabular*}{\columnwidth}{|c|*{6}{>{\centering\arraybackslash}p{1.2cm}|}}
        \hline
        \textbf{CASE} & \textbf{Accuracy (\%)} & \textbf{Sensitivity (\%)} & \textbf{Specificity (\%)} & \textbf{Precision (\%)} & \textbf{Recall (\%)} & \textbf{F-measure} \\
        \hline
        A-C & 72.00 & 82.65 & 61.52 & 67.89 & 82.65 & 0.74 \\
        \hline
        A-D & 72.50 & 97.03 & 47.76 & 65.55 & 97.03 & 0.78 \\
        \hline
        A-E & 100.00 & 100.00 & 100.00 & 100.00 & 100.00 & 1.00 \\
        \hline
        B-C & 82.00 & 67.22 & 95.48 & 95.60 & 67.22 & 0.78 \\
        \hline
        B-D & 68.00 & 91.43 & 45.27 & 62.48 & 91.42 & 0.74 \\
        \hline
        B-E & 99.50 & 99.23 & 100.00 & 100.00 & 99.23 & 0.99 \\
        \hline
    \end{tabular*}
    \captionsetup{justification=centering}
    
    \caption{Results of the proposed model for the combination of ICA with NB}
    \label{tab:tab4}
\end{table}

Tables \ref{tab:tab5}, \ref{tab:tab6} and \ref{tab:tab7} represent the various performance measures for the PCA dimensionality reduction technique with three different classifiers: NB, SVM and KNN. As per the table data, we have noticed that the highest accuracy, $100\%$, is obtained in the case of the A-E dataset for the NB classifier. We got the highest average recall value, and the F-measures for the A-E dataset is $100\%$. Specificity values of $100\%$ were achieved for the datasets A-E and B-E for SVM and KNN, respectively. 

\begin{table}[!htb]
    \centering
    \setlength{\tabcolsep}{0.157\tabcolsep}
    \begin{tabular*}{\columnwidth}{|c|*{6}{>{\centering\arraybackslash}p{1.2cm}|}}
        \hline
        \textbf{CASE} & \textbf{Accuracy (\%)} & \textbf{Sensitivity (\%)} & \textbf{Specificity (\%)} & \textbf{Precision (\%)} & \textbf{Recall (\%)} & \textbf{F-measure} \\
        \hline
        A-C & 81.00 & 88.47 & 71.87 & 78.36 & 88.47 & 0.83 \\
        \hline
        A-D & 90.50 & 93.20 & 89.28 & 89.34 & 93.20 & 0.91 \\
        \hline
        A-E & 58.00 & 100.00 & 18.18 & 57.18 & 100.00 & 0.71 \\
        \hline
        B-C & 81.00 & 82.32 & 77.93 & 79.39 & 82.32 & 0.81 \\
        \hline
        B-D & 83.50 & 83.09 & 81.94 & 83.12 & 83.09 & 0.89 \\
        \hline
        B-E & 88.50 & 100.00 & 75.07 & 86.68 & 100.00 & 0.92 \\
        \hline
    \end{tabular*}
    \captionsetup{justification=centering}
    
    \caption{Results of the proposed model PCA with KNN Algorithm}
    \label{tab:tab5}
\end{table}

\begin{table}[!htb]
    \centering
    \setlength{\tabcolsep}{0.157\tabcolsep}
    \begin{tabular*}{\columnwidth}{|c|*{6}{>{\centering\arraybackslash}p{1.2cm}|}}
        \hline
        \textbf{CASE} & \textbf{Accuracy (\%)} & \textbf{Sensitivity (\%)} & \textbf{Specificity (\%)} & \textbf{Precision (\%)} & \textbf{Recall (\%)} & \textbf{F-measure} \\
        \hline
        A-C & 77.50 & 96.87 & 55.57 & 71.09 & 96.87 & 0.81 \\
        \hline
        A-D & 84.50 & 97.07 & 71.43 & 77.77 & 97.07 & 0.86 \\
        \hline
        A-E & 93.50 & 100.00 & 86.87 & 89.34 & 100.00 & 0.94 \\
        \hline
        B-C & 83.00 & 93.65 & 71.89 & 77.89 & 93.65 & 0.85 \\
        \hline
        B-D & 85.00 & 93.85 & 74.36 & 80.27 & 93.85 & 0.86 \\
        \hline
        B-E & 90.00 & 100.00 & 80.09 & 83.98 & 100.00 & 0.91 \\
        \hline
    \end{tabular*}
    \captionsetup{justification=centering}
    
    \caption{Results of the proposed model PCA with SVM Algorithm}
    \label{tab:tab6}
\end{table}

\begin{table}[!htb]
    \centering
    \setlength{\tabcolsep}{0.157\tabcolsep}
    \begin{tabular*}{\columnwidth}{|c|*{6}{>{\centering\arraybackslash}p{1.2cm}|}}
        \hline
        \textbf{CASE} & \textbf{Accuracy (\%)} & \textbf{Sensitivity (\%)} & \textbf{Specificity (\%)} & \textbf{Precision (\%)} & \textbf{Recall (\%)} & \textbf{F-measure} \\
        \hline
        A-C & 80.50 & 94.45 & 67.12 & 74.17 & 94.45 & 0.83 \\
        \hline
        A-D & 80.00 & 96.26 & 63.67 & 72.64 & 96.26 & 0.82 \\
        \hline
        A-E & 100.00 & 100.00 & 100.00 & 100.00 & 100.00 & 1.00 \\
        \hline
        B-C & 90.50 & 84.28 & 95.71 & 97.50 & 84.28 & 0.90 \\
        \hline
        B-D & 89.00 & 90.31 & 88.38 & 87.95 & 90.31 & 0.89 \\
        \hline
        B-E & 99.50 & 99.00 & 100.00 & 100.00 & 99.00 & 0.99 \\
        \hline
    \end{tabular*}
    \captionsetup{justification=centering}
    
    \caption{Results of the proposed model PCA with NB Algorithm}
    \label{tab:tab7}
\end{table}

\begin{table}[!htb]
    \centering
    \setlength{\tabcolsep}{0.157\tabcolsep}
    \begin{tabular*}{\columnwidth}{|c|*{6}{>{\centering\arraybackslash}p{1.2cm}|}}
        \hline
        \textbf{CASE} & \textbf{Accuracy (\%)} & \textbf{Sensitivity (\%)} & \textbf{Specificity (\%)} & \textbf{Precision (\%)} & \textbf{Recall (\%)} & \textbf{F-measure} \\
        \hline
        A-C & 77.50 & 82.95 & 70.30 & 76.89 & 82.95 & 0.78 \\
        \hline
        A-D & 66.50 & 64.45 & 67.84 & 69.40 & 64.45 & 0.66 \\
        \hline
        A-E & 92.00 & 100.00 & 84.65 & 86.49 & 100.00 & 0.92 \\
        \hline
        B-C & 76.50 & 64.08 & 87.32 & 82.89 & 64.08 & 0.72 \\
        \hline
        B-D & 80.00 & 73.44 & 82.98 & 85.54 & 73.44 & 0.77 \\
        \hline
        B-E & 90.00 & 100.00 & 79.61 & 84.92 & 100.00 & 0.91 \\
        \hline
    \end{tabular*}
    \captionsetup{justification=centering}
    
    \caption{Results of proposed model LDA with KNN Algorithm}
    \label{tab:tab8}
\end{table}

\begin{table}[!htb]
    \centering
    \setlength{\tabcolsep}{0.157\tabcolsep}
    \begin{tabular*}{\columnwidth}{|c|*{6}{>{\centering\arraybackslash}p{1.2cm}|}}
        \hline
        \textbf{CASE} & \textbf{Accuracy (\%)} & \textbf{Sensitivity (\%)} & \textbf{Specificity (\%)} & \textbf{Precision (\%)} & \textbf{Recall (\%)} & \textbf{F-measure} \\
        \hline
        A-C & 100.00 & 100.00 & 100.00 & 100.00 & 100.00 & 1.00 \\
        \hline
        A-D & 72.00 & 72.48 & 73.36 & 72.75 & 72.48 & 0.71 \\
        \hline
        A-E & 96.00 & 99.09 & 90.63 & 95.56 & 99.09 & 0.97 \\
        \hline
        B-C & 91.00 & 86.70 & 94.02 & 93.57 & 86.70 & 0.90 \\
        \hline
        B-D & 100.00 & 100.00 & 100.00 & 100.00 & 100.00 & 1.00 \\
        \hline
        B-E & 76.00 & 88.38 & 63.88 & 74.31 & 88.38 & 0.80 \\
        \hline
    \end{tabular*}
    \captionsetup{justification=centering}
    
    \caption{Results of proposed model LDA with SVM Algorithm}
    \label{tab:tab9}
\end{table}

\begin{table}[!htb]
    \centering
    \setlength{\tabcolsep}{0.157\tabcolsep}
    \begin{tabular*}{\columnwidth}{|c|*{6}{>{\centering\arraybackslash}p{1.2cm}|}}
        \hline
        \textbf{CASE} & \textbf{Accuracy (\%)} & \textbf{Sensitivity (\%)} & \textbf{Specificity (\%)} & \textbf{Precision (\%)} & \textbf{Recall (\%)} & \textbf{F-measure} \\
        \hline
        A-C & 100.00 & 100.00 & 100.00 & 100.00 & 100.00 & 1.00 \\
        \hline
        A-D & 100.00 & 100.00 & 100.00 & 100.00 & 100.00 & 1.00 \\
        \hline
        A-E & 100.00 & 100.00 & 100.00 & 100.00 & 100.00 & 1.00 \\
        \hline
        B-C & 100.00 & 100.00 & 100.00 & 100.00 & 100.00 & 1.00 \\
        \hline
        B-D & 100.00 & 100.00 & 100.00 & 100.00 & 100.00 & 1.00 \\
        \hline
        B-E & 100.00 & 100.00 & 100.00 & 100.00 & 100.00 & 1.00 \\
        \hline
    \end{tabular*}
    \captionsetup{justification=centering}
    
    \caption{Results of proposed model LDA with NB Algorithm}
    \label{tab:tab10}
\end{table}

Tables \ref{tab:tab8}, \ref{tab:tab9} and \ref{tab:tab10} represent the various performance measures owing to the NB, SVM and KNN classifiers using the LDA dimensionality reduction technique. The comparison of Tables \ref{tab:tab2}, \ref{tab:tab3}, \ref{tab:tab4}, \ref{tab:tab5}, \ref{tab:tab6}, \ref{tab:tab7}, \ref{tab:tab8}, \ref{tab:tab9} and \ref{tab:tab10} shows that the combination (LDA+NB) provides the best result for all the measures and all the data sets. The NB classifier achieved an accuracy of $100\%$ for the dataset: A-C, A-D, A-E, B-C, B-D, and B-E. Specificity and precision values of $100\%$ are achieved for all the dataset combinations (A-C, A-D, A-E, B-C, B-D, and B-E).   

\begin{figure}[!htb]
    \centering
    \includegraphics[width=0.4\textwidth]{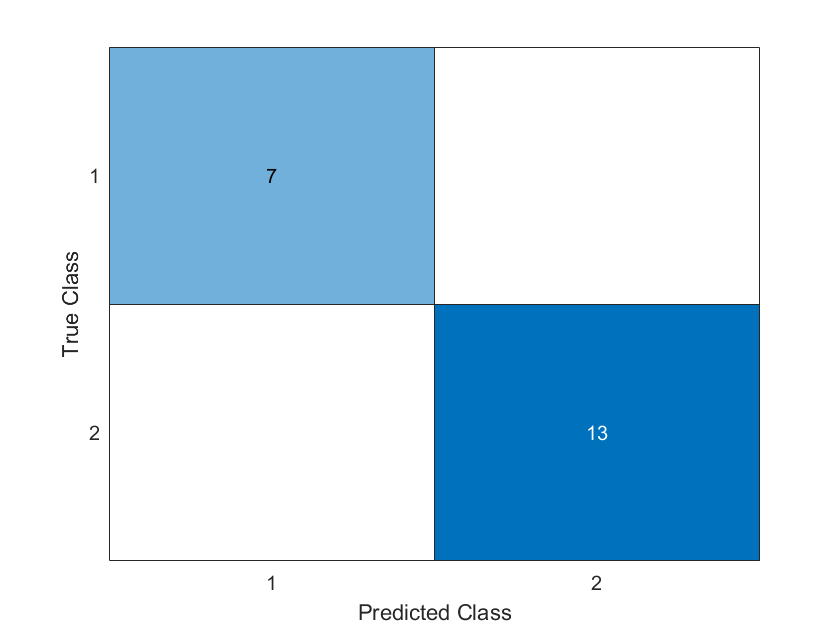}
    \captionsetup{justification=centering}
    \caption{Confusion matrix for PCA with KNN Algorithm}
    \label{fig:fig5}
\end{figure}

\begin{figure}[!htb]
    \centering
    \includegraphics[width=0.4\textwidth]{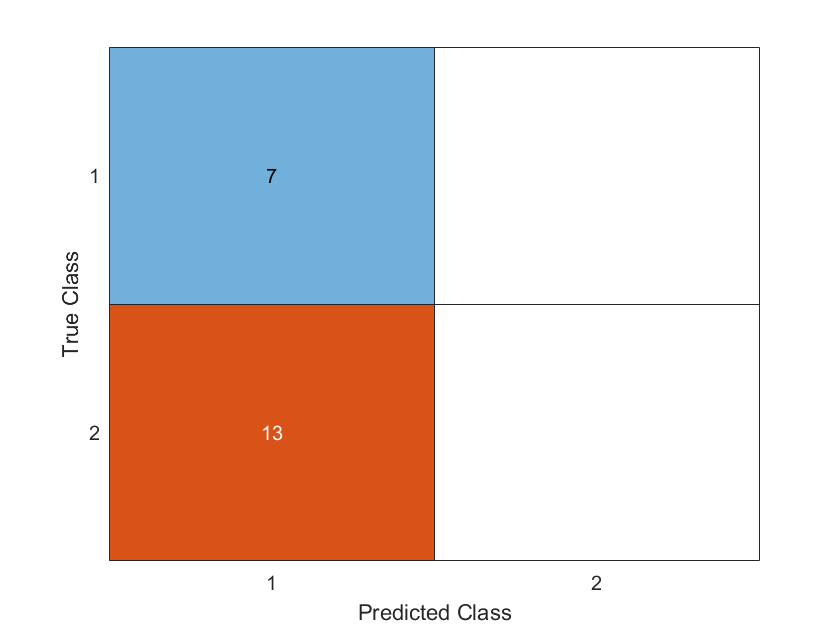}
    \captionsetup{justification=centering}
    \caption{Confusion matrix for PCA with SVM Algorithm}
    \label{fig:fig6}
\end{figure}

\begin{figure}[!htb]
    \centering
    \includegraphics[width=0.4\textwidth]{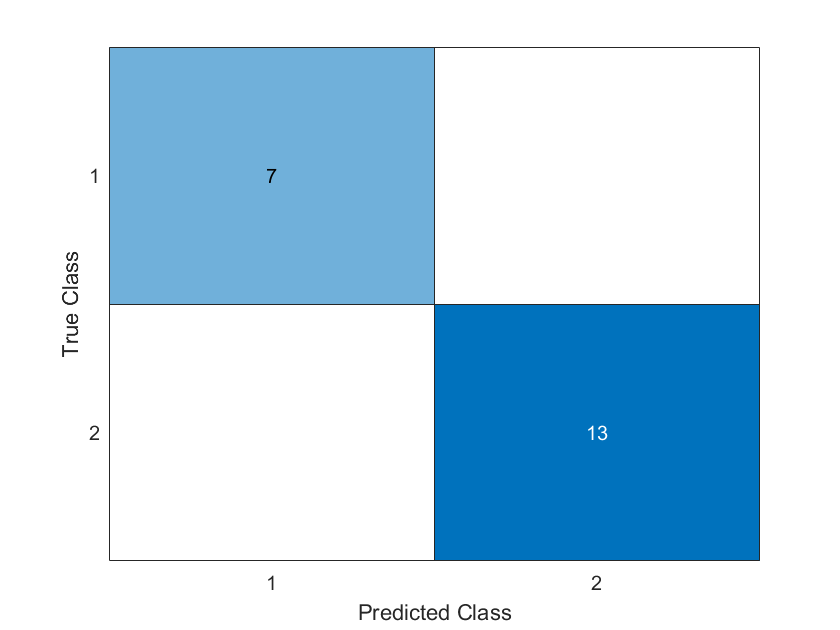}
    \captionsetup{justification=centering}
    \caption{Confusion matrix for PCA with NB Algorithm}
    \label{fig:fig7}
\end{figure}

In Figures \ref{fig:fig5}, \ref{fig:fig6} and \ref{fig:fig7}, the confusion matrix shows two classes, e.g., the true and predicted classes. It shows the confusion matrix true and predicted classes for PCA with three different classifier algorithms. There are different matrices for SVM, but other classifiers have the same matrices. It shows two classes, e.g., the true and predicted classes. 

\begin{figure}[!htb]
    \centering
    \includegraphics[width=0.4\textwidth]{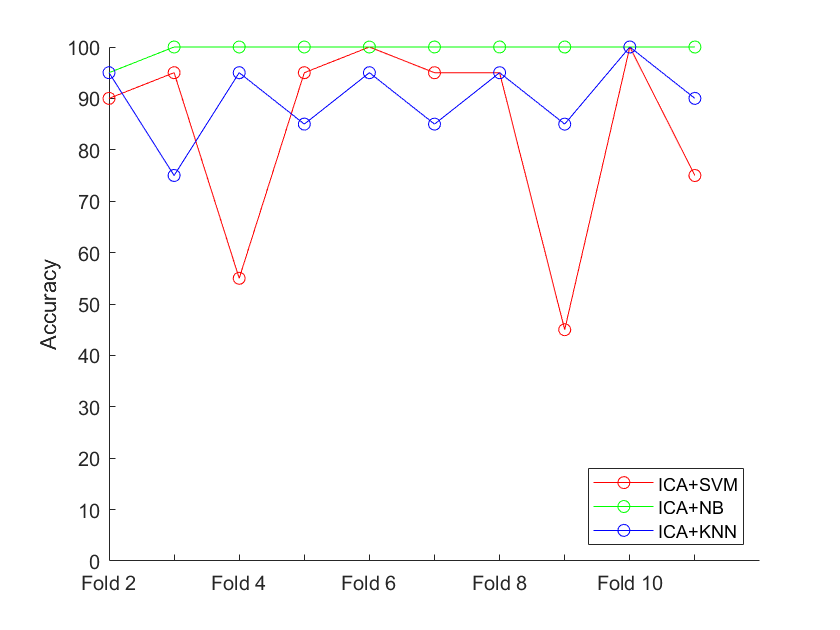}
    \captionsetup{justification=centering}
    \caption{Fold-wise accuracy using ICA and SVM, NB, KNN}
    \label{fig:fig8}
\end{figure}

\begin{figure}[!htb]
    \centering
    \includegraphics[width=0.4\textwidth]{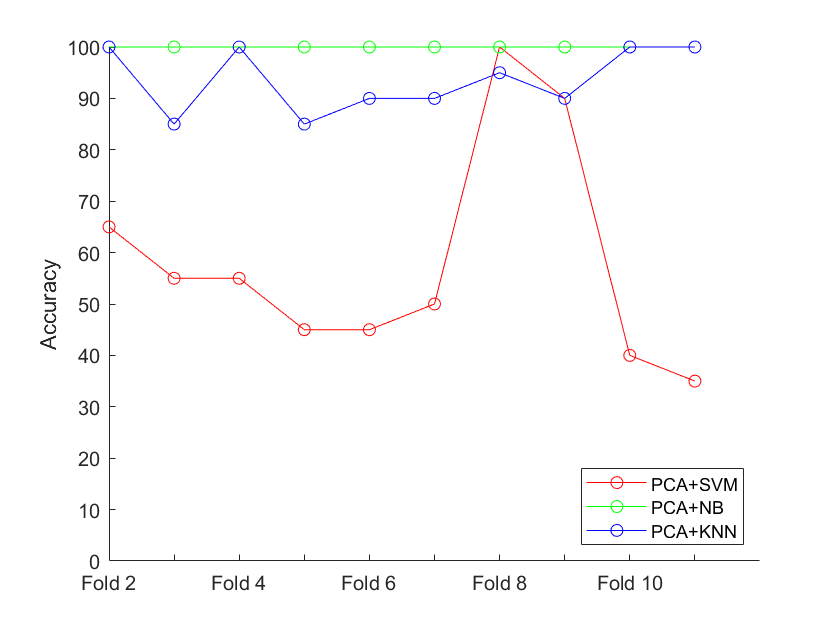}
    \captionsetup{justification=centering}
    \caption{Fold-wise accuracy using PCA and SVM, NB, KNN}
    \label{fig:fig9}
\end{figure}

\begin{figure}[!htb]
    \centering
    \includegraphics[width=0.4\textwidth]{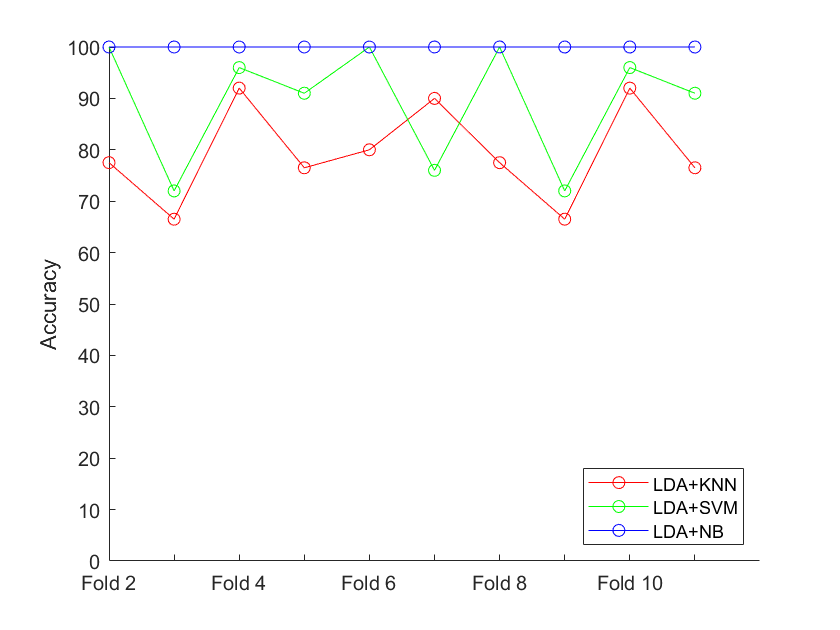}
    \captionsetup{justification=centering}
    \caption{Fold-wise accuracy using LDA and SVM, NB, KNN}
    \label{fig:fig10}
\end{figure}

Fig. \ref{fig:fig8}, \ref{fig:fig9} and \ref{fig:fig10} shows the average accuracy for different folds in the case of ICA, PCA and LDA for the three different classifiers: SVM, NB, and KNN. From Figure \ref{fig:fig8}, it is observed that (ICA+SVM) combination provides a better result compared to (ICA+NB) and (ICA+KNN). Fig. \ref{fig:fig9} shows that the SVM classifier offers a better result than KNN and NB for the case of PCA. But for LDA, it is noticed from Fig.\ref{fig:fig10} that (LDA+NB) combination provides a better result compared to the (LDA+KNN) and (LDA+SVM) combinations.

\begin{figure}[!htb]
    \centering
    \includegraphics[width=0.4\textwidth]{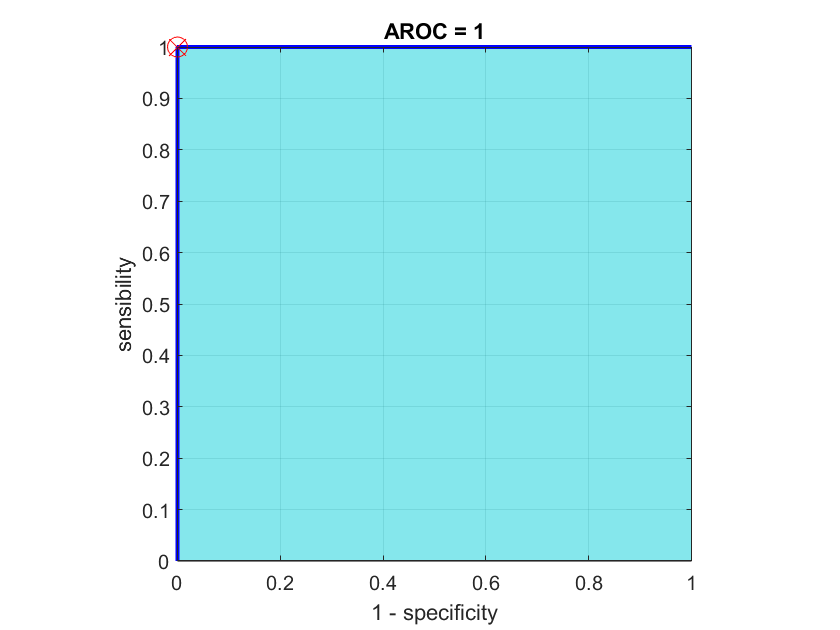}
    \captionsetup{justification=centering}
    \caption{ROC plot for PCA, ICA, LDA and SVM, NB, KNN}
    \label{fig:fig11}
\end{figure}

Fig. \ref{fig:fig11} shows the ROC plots for PCA, ICA, LDA and SVM, NB, and KNN. All plots are the same because almost all the results reached approximately $100$. So, the same curve for all. Three classifiers obtained ROC plots from the dataset of $5$ different sets of EEG signals.

\pagebreak

\section{Conclusion} \label{sec:conclusion}

EEG signals are widely used for the detection and diagnosis of epileptic seizures. Clinical diagnosis of epilepsy is a time-killing and costly process. So, it is required to design a simple and fast technique for classifying EEG signals. In this work, a new epilepsy detection method has been presented by integrating DWT with three different dimensionality reduction techniques: PCA, LDA and ICA. In the following step, feature dimensions are further reduced by applying the feature-level fusion technique. Our proposed method is used for detecting epilepsy seizures. This combination of LDA with the NB method provides an accuracy of $100\%$, outperforming all existing methods. The dimension of the feature space of our proposed method is very low, so it would be a practical technique for treating epilepsy. In our future work, we will use larger datasets, such as the CMBH MIT dataset, to assess these epilepsy detection methods.

\bibliographystyle{IEEEtran}
\bibliography{references}

\vspace{-10mm}

\begin{IEEEbiography}[{\includegraphics[width=1in,height=1.25in,clip,keepaspectratio]{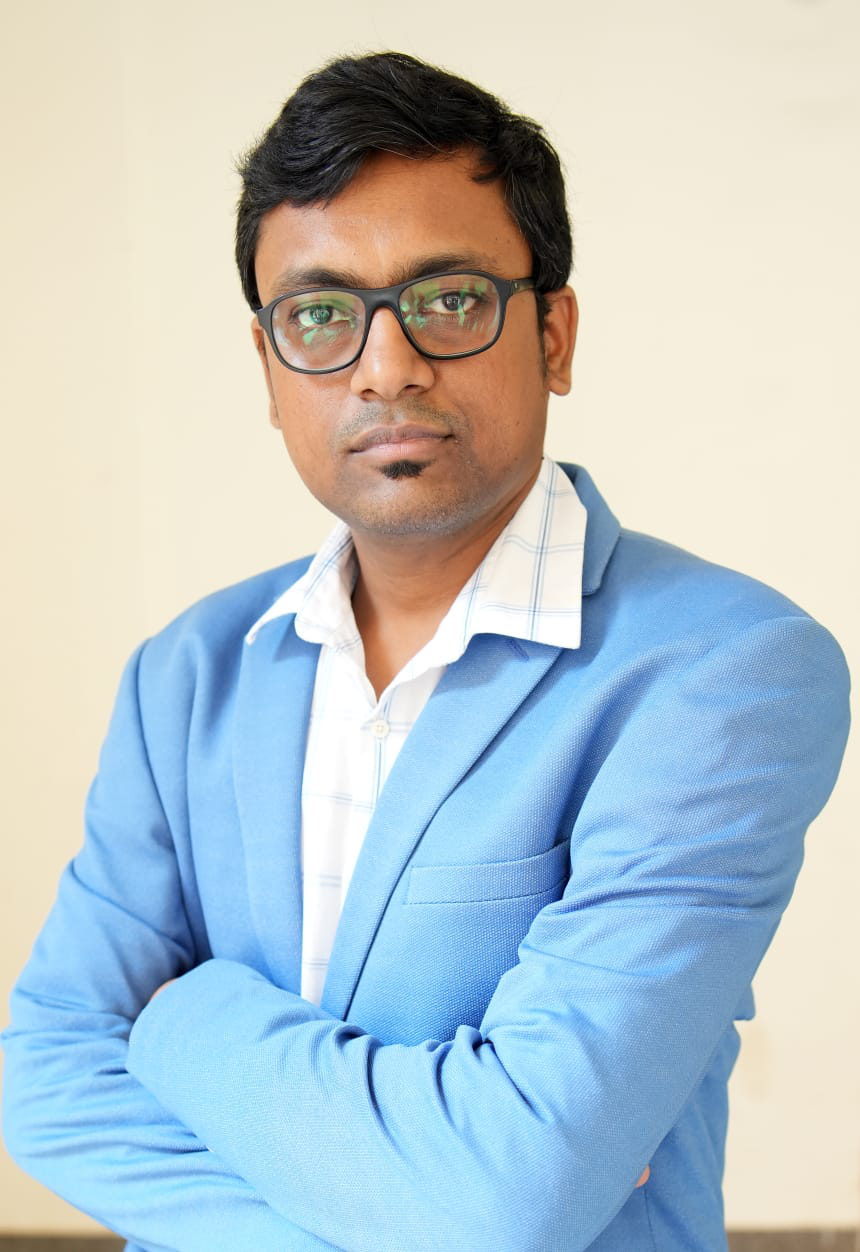}}]{RABEL GUHAROY} (Member IEEE) received the B.Tech degree in Information Technology from WBUT, Kolkata, India in 2012 and the Master of Technology degree in Computer Science and Engineering from WBUT, Kolkata, India in 2014 with GATE Scholarship. He is currently registered the Ph.D. degree in computer science and Engineering research from the National Institute of Technology, Durgapur, India.
He has extensive teaching experience and has 10 years experience in teaching. Now, Working as a Assistant Professor at Rasthriya Raksha University, SITAICS, Gandhinagar, India. It is a central government institute in India, under Ministry of Home Affairs.
\end{IEEEbiography}

\vspace{-7mm}

\begin{IEEEbiography}[{\includegraphics[width=1in,height=1.25in,clip,keepaspectratio]{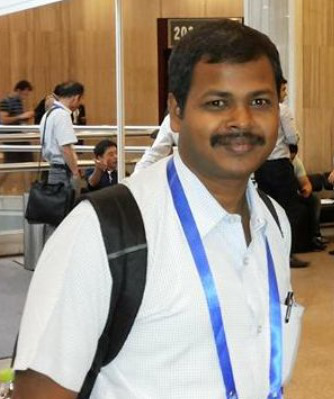}}]{NANDA DULAL JANA} received his B.Tech. and M.Tech. degree in computer science and engineering from the University of Calcutta, Kolkata, India, in 2004 and 2006, respectively, and the Ph.D. degree in computer science and engineering from the Indian Institute of Engineering Science and Technology, Shibpur, India, in 2017.
\end{IEEEbiography}

\vspace{-7mm}

\begin{IEEEbiography}[{\includegraphics[width=1in,height=1.25in,clip,keepaspectratio]{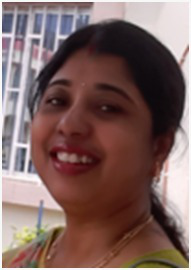}}]{SUPARNA BISWAS} completed her Ph.D. in Engineering from IIEST, Shibpur, Kolkata, India. She is presently working as Associate Professor in the Department of Electronics \& Communication Engineering in Guru Nanak Institute of Technology, Sodepur, Kolkata, India.
Her areas of research include image processing, machine learning and biomedical signal Processing.
\end{IEEEbiography}

\vspace{-7mm}

\begin{IEEEbiography}[{\includegraphics[width=1in,height=1.25in,clip,keepaspectratio]{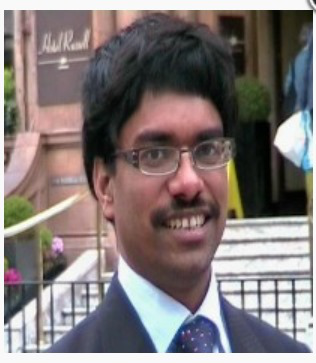}}]{LALIT GARG} completed her PhD in Engineering from the University of Ulster, UK. He is presently working as an Associate Professor in the Department of Computer Information Systems Faculty of Information \& Communication Technology, University of Malta, Malta.
Her areas of research include image processing, machine learning and biomedical signal Processing.
\end{IEEEbiography}

\EOD

\end{document}